\documentclass{article}
\usepackage{ijcai22}

\usepackage{times}
\usepackage{soul}
\usepackage{url}
\usepackage[hidelinks]{hyperref}
\usepackage[utf8]{inputenc}
\usepackage[small]{caption}
\usepackage{graphicx}
\usepackage{amsmath}
\usepackage{amsthm}
\usepackage{booktabs}
\usepackage{caption} 
\usepackage{graphicx}
\usepackage{booktabs} 
\usepackage{amssymb}
\usepackage{amsmath}
\usepackage[ruled, vlined, linesnumbered]{algorithm2e}
\usepackage{algorithmic}
\usepackage{multirow}
\usepackage{subcaption}
\usepackage{float}
\usepackage{mathtools}
\usepackage{bbold}
\usepackage{pgfplots}
\newcommand{\angstrom}{\mbox{\normalfont\AA}}
\usepackage{float}

\newcommand{\rulesep}{\unskip\ \vrule\ }

\title{Geometric Transformer for End-to-End Molecule Properties Prediction}

\author{
Yoni Choukroun\and
Lior Wolf
\affiliations
School of Computer Science, Tel Aviv University\\
\emails
    choukroun.yoni@gmail.com, wolf@cs.tau.ac.il
}

\begin{document}

\maketitle

\begin{abstract}
Transformers have become methods of choice in many applications thanks to their ability to represent complex interaction between elements. 
However, extending the Transformer architecture to non-sequential data such as molecules and enabling its training on small datasets remain a challenge. 
In this work, we introduce a Transformer-based architecture for molecule property prediction, which is able to capture the geometry of the molecule. 
We modify the classical positional encoder by an initial encoding of the molecule geometry, as well as a learned gated self-attention mechanism. 
We further suggest an augmentation scheme for molecular data capable of avoiding the overfitting induced by the overparameterized architecture. 
The proposed framework outperforms the state-of-the-art methods while being based on pure machine learning solely, i.e. the method does not incorporate domain knowledge from quantum chemistry and does not use extended geometric inputs beside the pairwise atomic distances.
\end{abstract}
\section{Introduction}

Properties of chemical compounds can generally be estimated using methods such as density functional theory (DFT) or \emph{ab initio} quantum chemistry \cite{jensen2017introduction}.
However, these can be computationally expensive and therefore have a limited applicability, especially for larger systems. 
In recent years, many approaches have started leveraging machine learning to reduce the computational complexity
required for efficiently predicting molecular properties.

In this vein, many contributions have focused on the creation of handcrafted representations at the atomic or molecular level \cite{christensen2020fchl,huang2016communication} as input for various machine learning methods. 
Schr\"{o}dinger's equation indicates that  the system variables that define the ground-state properties of a given molecule are a function of the \emph{inter-atomic distances} and the \emph{nuclear charges} solely. \cite{jensen2017introduction}. 
Based on this observation, several recent methods predict molecular properties in an end-to-end fashion where the input is defined by the  atoms' type and spatial position. Such methods often incorporate quantum chemistry knowledge and rely on extensive hyper-parameter tuning.

Since atomic interactions are challenging to simulate, many recent works make use of graph neural networks as a natural tool to model molecules \cite{gilmer2017neural}. 
Related to graph neural networks, Transformers \cite{vaswani2017attention} have recently become extremely popular in numerous application domains.

In this paper we extend the ubiquitous Transformer to chemical compounds data in order to predict their ground-state properties. Our work does not employ extended domain knowledge, and is based solely on the simple \emph{distance relevance assumption}, namely, that
{the bigger the distance between atomic elements, the lower the interaction}. 

Contrary to other works, the framework does not assume any extended input, e.g. quantum mechanical properties\cite{qiao2020orbnet}, complex geometric constants such as bending or torsion angles \cite{klicpera2020directional},
additional solvers, e.g. fast DFT as residual solvers (i.e. delta learning) \cite{unke2019physnet,qiao2020orbnet}), 
or even knowledge adaptation from quantum chemistry into the machine learning model design \cite{schutt2018schnet,klicpera2020directional}.

The Transformer we design is endowed with an adapted positional encoder, and with learned inter-atomic geometric embedding at the different levels of the model, allowing increased representational power, while maintaining the molecule invariance to rigid transformations and permutation. For better regularization, we suggest to augment the training set by merging pairs of molecules positioned far apart.

\begin{figure}[t]
\centering
\includegraphics[width=1\columnwidth]{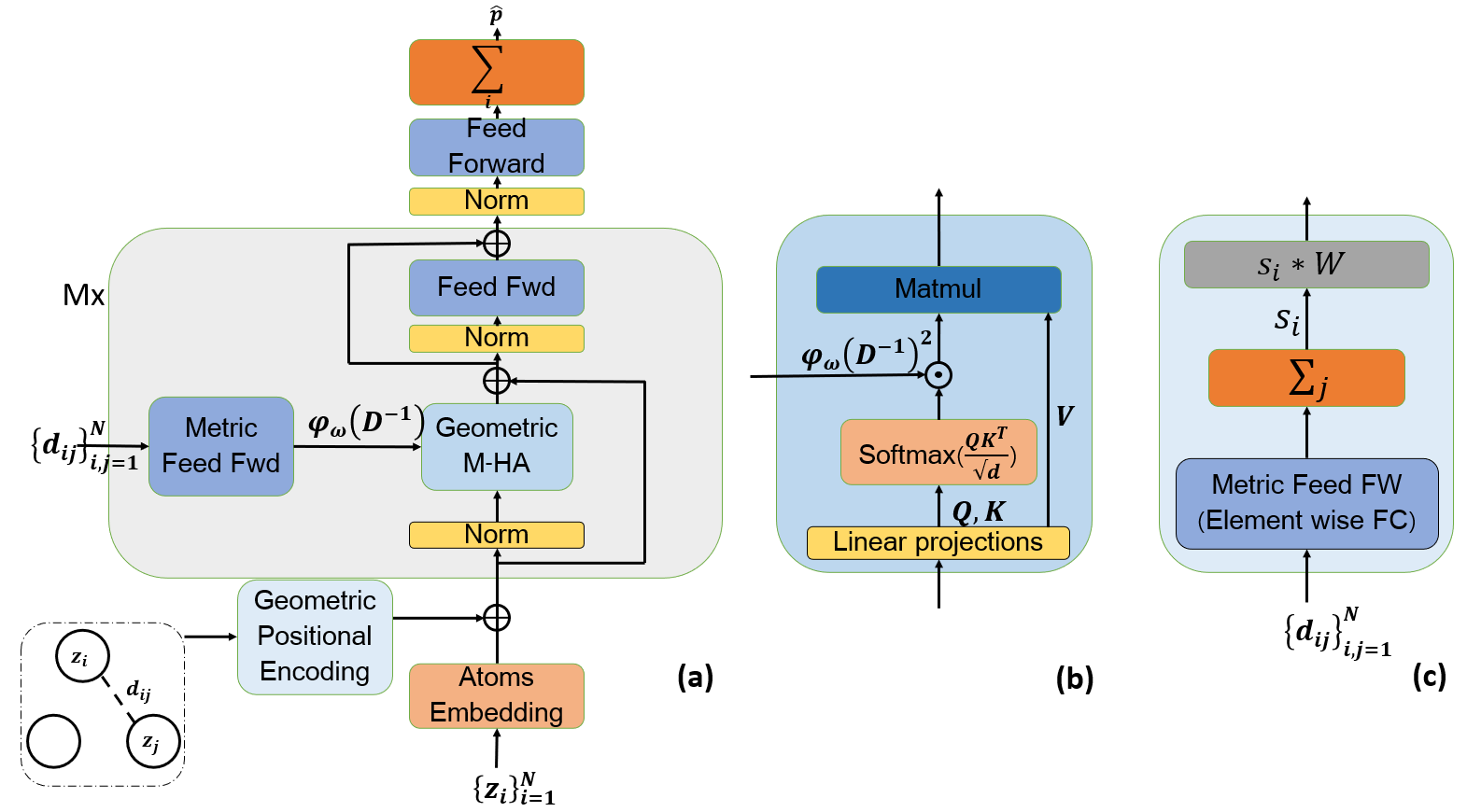}
\caption{The proposed Transformer architecture (a). 
The main architectural modifications are the initial positional encoding based on the pairwise distance matrix (c), and the metric learning module coupled with the augmented self-attention module (b). 
The model is composed of $M$ encoding blocks and the output block is composed of a normalized feed-forward neural network followed by a per atom $i$ summation to accumulate the contribution of each atom.}
\label{fig:model}
\vspace{-1.5em}
\end{figure}

The experimental results demonstrate the representational power of the \emph{end to end model},  potentially allowing removal of undesirable {inductive bias} \cite{goyal2020inductive} such as handcrafted envelope functions. Also, by allowing the model to fit the data while minimizing the domain-knowledge, scientific insights can emerge, {such as effective atomic cut-off radius, or most relevant molecular interactions for a given property. 
}
Our paper’s main contributions are:
\begin{itemize}
\item An initial positional encoding based on a learned mapping of the global embedding at the atomic level, which is able to encode the molecule geometry.

\item A new self-attention module where the positional encoding is inserted as an invariant \emph{gating}, enabling better realisation of the inter-atomic assumption. 

\item This self-attention mechanism is coupled with a learned inter-atomic pairwise \emph{metric} in order to learn adaptively the  molecule's soft adjacency matrix.

\item An augmentation technique for molecular data based on rigid transformations. This regularization approach can be used for permutation-invariant models where other regularization techniques such as dropout cannot be applied in a straightforward manner.

\item A novel end-to-end Transformer architecture setting state-of-the-art performance, outperforming all other Transformer based methods by a large margin. In contrast to existing methods, the proposed approach does not require any external information or knowledge, directly demonstrating the power and flexibility of suitably designed Transformers.

\end{itemize}

\section{Related Work}
Classical molecule property prediction methods combined handcrafted features generally based on force field methods at the atom level \cite{christensen2020fchl} or molecular level  \cite{huang2016communication}, integrated into various machine learning models such as Kernel methods \cite{christensen2020fchl}
, Gaussian processes 
\cite{bartok2010gaussian}
or Neural networks 
\cite{schutt2018schnet}.

These methods have recently been superseded by end-to-end neural networks, alleviating the need for handcrafted signatures.
The most popular and powerful models are based on graph neural networks, which allow a natural representation of the molecular graph \cite{schutt2018schnet,unke2019physnet,anderson2019cormorant,klicpera2020directional,qiao2020orbnet}.
These architectures are generally designed based on the message-passing mechanism, by aggregating features obtained from atom types, geometric invariants such as pairwise inter-atomic  distances \cite{schutt2018schnet,unke2019physnet}, bending or torsion angles \cite{klicpera2020directional}, or handcrafted atomic features derived from quantum mechanics \cite{qiao2020orbnet}.
Existing methods generally involve a deep understanding and a cautious adaptation of the underlying physics in order to provide better preconditioning
\cite{klicpera2020directional,qiao2020orbnet}.

Transformer neural networks were originally introduced for machine translation \cite{vaswani2017attention} and they now dominate most applications in the field of Natural Language Processing. 
Transformer encoders primarily rely on the self-attention operation in conjunction with feed-forward layers, allowing manipulation of variable-size sequences and learning of long-range dependencies.
Many works have augmented the self-attention mechanism using domain-specific knowledge \cite{chen2017neural,bello2019attention}.

Recently, a transformer architecture called MAT has been proposed for chemical molecules \cite{maziarka2020molecule,maziarka2021relative}.
MAT modifies the self-attention module by summing the inverse exponent of the pairwise distance matrix to the self-attention tensor.
As one contribution of our work, we show that the MAT self-attention architecture is sub-optimal in modelling interactions, and we propose a better self-attention module capable of capturing the connectivity of the graph.  Our method also outperforms concurrent efforts \cite{wu20213d,kwak2021geometry} which integrate mollifiers from the literature \cite{schutt2018schnet,klicpera2020directional} to the \emph{key} element of the self-attention.    

\section{Method}
A molecule is defined by the atomic numbers
\mbox{$z = \{z_1, \dots , z_N \} \in \mathbb{Z}_{+}$}, which serve to identify each type of atom,  and the three dimensional positions $X = \{x_1, \dots , x_N \} \in \mathbb{R}^3$ of the $N$ atoms composing it.
Molecular predictions must satisfy fundamental symmetries and invariance of physical laws such as invariance to rigid spatial transformation (rotation and translation) and permutation (atoms of the same type are indistinguishable).
Therefore, the positional input is transformed to interatomic Euclidean distances $D=\{d_{ij}\}_{i,j=1}^{N}$ where $d_{ij}=\|x_i-x_j\|_2$ for rigid transform invariance, while permutation invariance is obtained via equal initial atomic representations of identical particles.

In this work, we design a parameterized deep neural network $f_{\theta}$ for scalar regression of properties $p\in \mathbb{R}$ such that $f_{\theta} : \{z, D\} \rightarrow \mathbb{R}$.  
We do not include any further auxiliary features in the input (e.g. adjacency, bonds type, hybridization \cite{gilmer2017neural,maziarka2020molecule}, DFT solver \cite{unke2019physnet,qiao2020orbnet}), neither any kind of knowledge adaptation from quantum physics/chemistry \cite{unke2019physnet,klicpera2020directional,klicpera2020fast,qiao2020orbnet}.
Figure \ref{fig:model} depicts the proposed architecture. 
The positional encoding provides an initial geometry aware embedding of the atoms while the self-attention mechanism enables
the accurate learning of the molecule geometry as well as the
determination of the complex geometric interactions that are
modeled in order to perform the regression task.


\subsection{Transformer}
Transformer was introduced by \cite{vaswani2017attention} as a novel, attention-based building block for machine translation. The input sequence is first embedded into a high-dimensional space, coupled with positional embedding for each element. The embeddings are then propagated through multiple normalized self-attention and feed-forward blocks. 

The self-attention mechanism introduced by Transformers is based on a trainable associative memory with (key, value) vector pairs where a query vector $q \in \mathbb{R}^d$ is matched against a set of $k$ key vectors using scaled inner products as follows
\begin{equation}
\begin{aligned}
\label{transformer_att}
  \vspace{-0.5em}
A(Q,K,V)=\text{Softmax}\bigg(\frac{QK^{T}}{\sqrt{d}}\bigg)V,
\end{aligned}
\end{equation}
where $Q \in \mathbb{R}^{N \times d}$, $K \in \mathbb{R}^{k \times d}$ and $V \in \mathbb{R}^{k \times d}$ represent the packed $N$ queries, $k$ keys and values tensors respectively.
Keys, queries and values are obtained using linear transformations of the sequence' elements.
A multi-head self-attention layer is defined by extending the self-attention using $h$ attention \emph{heads}, i.e. $h$ self-attention functions applied to the input, reprojected to values via a $dh \times D$ linear layer.
\begin{figure}[t!]
\centering
  \includegraphics[trim={35 100 31 80},clip, width=0.47\columnwidth]{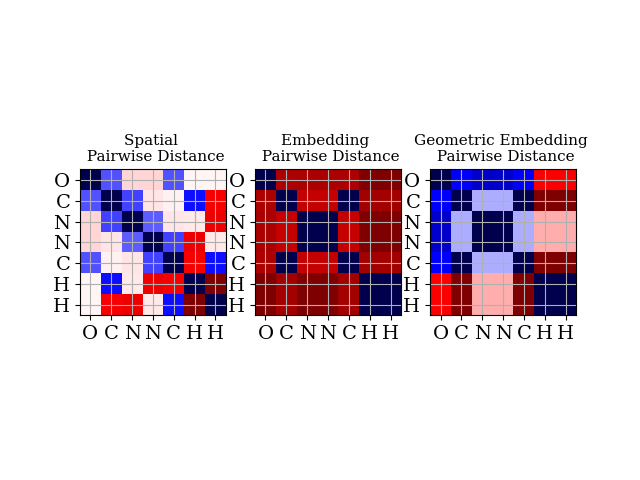}
\rulesep
  \includegraphics[trim={35 100 31 80},clip, width=0.47\columnwidth]{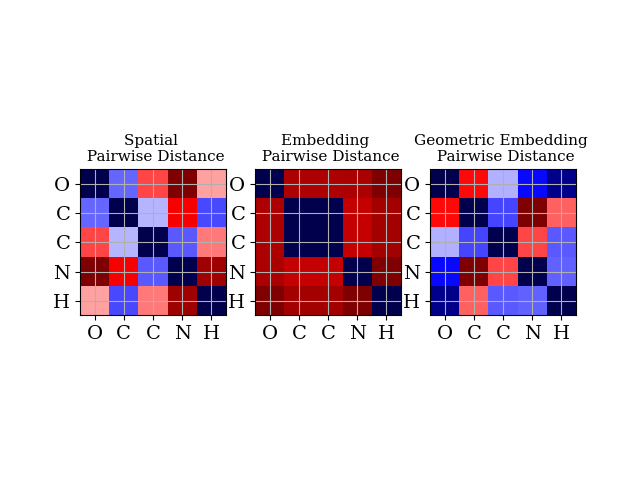}
\caption{The impact of the initial positional encoder on the embedding for two different molecules (left and right).
We show the pairwise distance matrix $D$ (left), pairwise distance of the initial embedding $Emb(z)$ (middle), and the pairwise distance of the final geometric embedding (right).
The marks on the map represent the type of the atoms $z_{i}$.
Cold and warm colors represent low and high values respectively.
}
\label{fig:emb}
\end{figure}
\begin{figure}[t!]
  \begin{minipage}[c]{0.50\columnwidth}
  \centering
    \includegraphics[trim={0 10 10 0},clip, width=1\columnwidth]{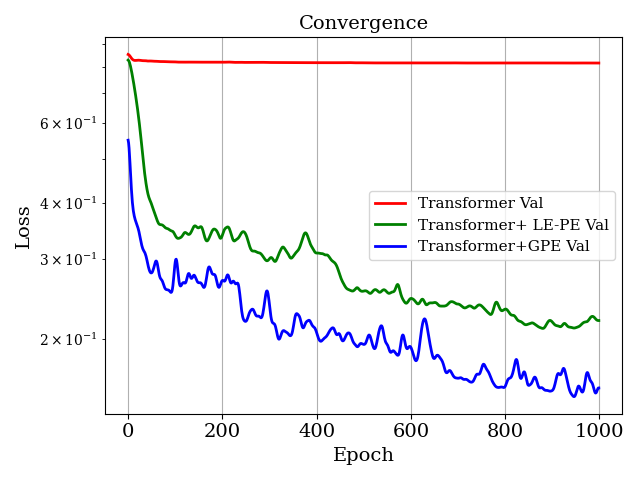}
  \end{minipage}\hfill
    \begin{minipage}[c]{0.45\columnwidth}
\caption{
The validation $L_{1}$ loss (U0 property) of a regular transformer without any geometric input (red), the same transformer with our Laplacian extension
(green), and with the proposed initial geometric positional encoding (GPE).
}
\label{fig:emb2}
  \end{minipage}\hfill
\vspace{-1.5em}
\end{figure}

\subsection{Geometric Positional Encoding}
The initial embedding of the molecule is based solely on the atoms' type and thus is unable to differentiate similar atoms {since the molecule geometry is omitted.}
The original Transformer's positional encoding module aims to transfer a measure of proximity of the sequence elements to the initial embedding.
In our case, since the input is defined as a set rather than a sequence, the  positional encoder needs to be adapted in order to provide a geometry-aware initial embedding.
Here we propose to use the pairwise inter-atomic distance matrix in order to bring positional information to each atom.
For each atom we first embed its pairwise distance vector to a single scalar in order to keep the module invariant to the size of the molecule, and then project it to the initial embedding space.
Formally, denoting the atom embedding $Emb(z_{i}): \mathbb{Z}_{+}\rightarrow \mathbb{R}^{d}$ and the pairwise distance matrix \mbox{$D \in \mathbb{R^{+}}^{n\times n}$} such that $(D)_{ij}:=(D_{i})_{j}=d_{ij}$, we have 
\begin{equation}
\begin{aligned}
\label{eigenmap}
y_{i} = Emb(z_{i}) + W\sum_{j}f_{\text{pos}}(d_{ij}).
\end{aligned}
\vspace{-0.5em}
\end{equation}
Here, $y_{i}$ denotes the obtained initial positioned embedding of atom $z_{i}$, $f_{\text{pos}}:\mathbb{R^{+}}\rightarrow \mathbb{R}$ denotes the mapping of the pairwise distance {parameterized as a shallow neural network} , and $W\in \mathbb{R}^{d}$ denotes the projection matrix of the atomic one-dimensional embedding onto the embedding space.
Figure \ref{fig:emb} depicts the impact of the positional encoder on the initial embedding.
As can be observed, the initial embedding does not differentiate between atoms of the same type, while positional encoding brings information about the geometry.
The convergence plot in Figure \ref{fig:emb2} demonstrates that the proposed positional encoding allows the transformer to learn the molecular geometry and thus be able to predict properties of the molecule.
We also compare our method with the positional encoder of \cite{dwivedi2020generalization} where, since in our setting no graph is given, we propose to compute the 15 (half the biggest molecule size) first Laplacian eigenmaps \cite{belkin2003laplacian} instead of the combinatorial Laplacian. 
Our approach allows to encode the global geometry of the molecule with respect to each atom, while the permutation-invariant aggregation part in Eq. (\ref{eigenmap}) maintains the symmetry invariance of the embedding.
{This preconditioning enables faster convergence and slightly better performance.}


\begin{figure*}[h]
\centering
\begin{subfigure}{.19\textwidth}
  \centering
  \includegraphics[width=1\linewidth]{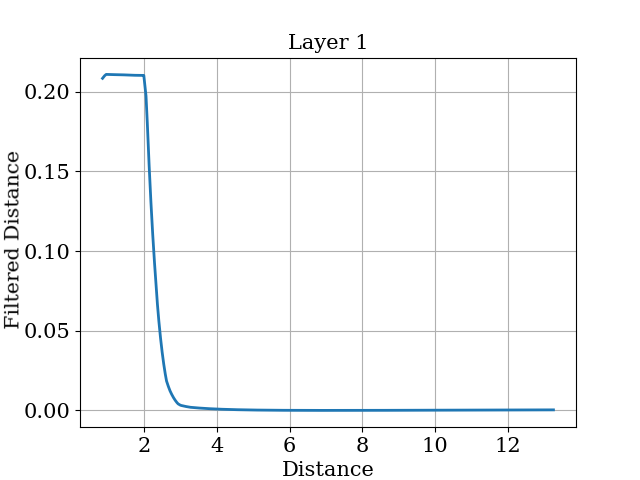}
\end{subfigure}%
\hspace{0em}
\begin{subfigure}{.19\textwidth}
  \centering
  \includegraphics[width=1\linewidth]{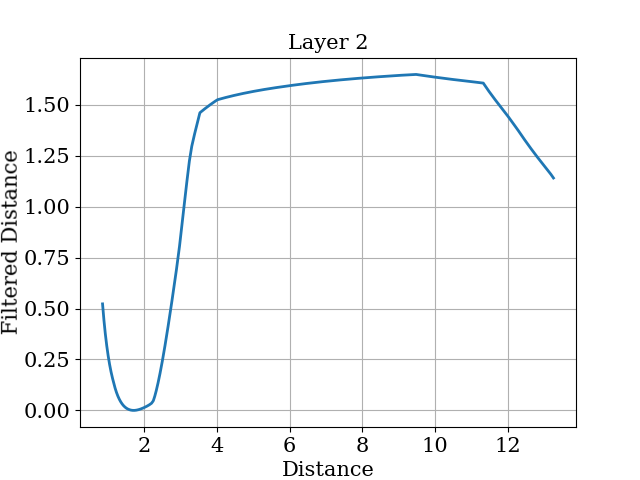}
\end{subfigure}%
\hspace{0em}
\begin{subfigure}{.19\textwidth}
  \centering
  \includegraphics[width=1\linewidth]{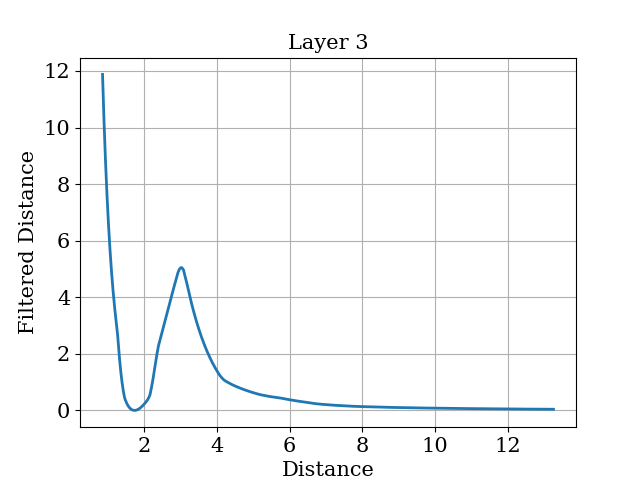}
\end{subfigure}
\begin{subfigure}{.19\textwidth}
  \centering
  \includegraphics[width=1\linewidth]{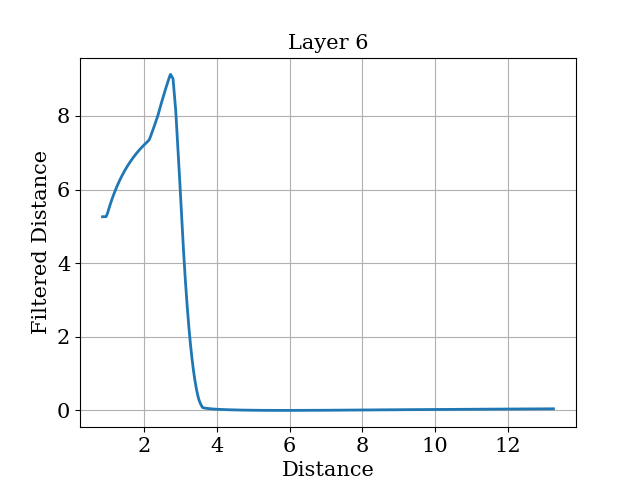}
\end{subfigure}
\hspace{0em}
\begin{subfigure}{.19\textwidth}
  \centering
  \includegraphics[width=1\linewidth]{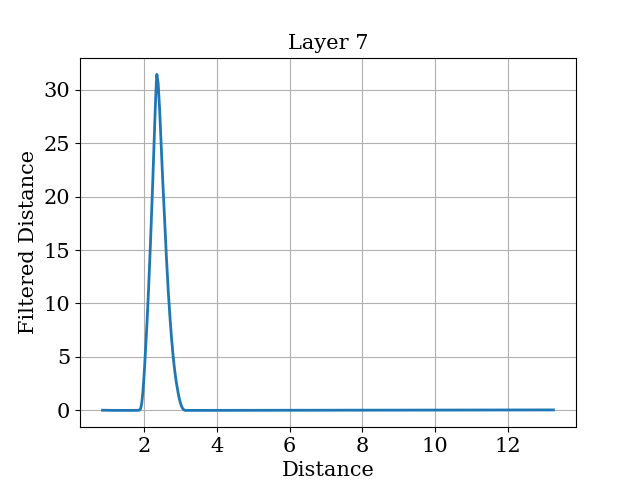}
\end{subfigure}
  \vspace{-0.5em}
\caption{1D cuts of the learned spherical metric at different blocks of the proposed Transformer. The distance is in $\angstrom$.
One can observe different low-, band- and high-pass filters, which demonstrates the model's ability to learn the connectivity that is  relevant for each level of the network. 
Notice the different dynamic ranges of the filtered values, which lead to different levels of impact on the self-similarity matrix.}
\vspace{-1.5em}
\label{fig:filters}
\end{figure*}

\subsection{Geometric Self-Attention}
\label{sec:geo_self_att}
In order to maintain the invariant properties of molecules, we propose to augment the initial positional encoding layer and import the pairwise information directly into the self-attention layer. 
This is a natural choice since the self-attention layer already computes pairwise similarity of the atoms' representations via the normalized inner product.
The self-attention layer (Eq. \ref{transformer_att}) is then extended to
\begin{equation}
\begin{aligned}
\label{transformer_att_2}
\Big(\tilde{A}(Q,K,V,D)\Big)_{i}&=\varphi_{\omega}(Q_{i},K,D_{i})V \\
&=\sum_{j=1}^{k}\varphi_{\omega}(Q_{i},K_j,D_{ij})V_{j},
\end{aligned}
\vspace{-0.5em}
\end{equation}
where $D$ denote the pairwise distance matrix, and $\varphi_{\omega}$ a  parameterized, potentially learned, similarity function. 
Several formulations can be conceived, however every extension should satisfy the \emph{distance relevance assumption} such that $\varphi_{\omega}$ is a vanishing mapping with respect to distance $D_{ij}$, such that for a given positive $\delta$ scalar for all $D_{ij}>\delta$ we would have $\varphi_{\omega}(Q_{i},K_j,D_{ij})\to 0.$

These assumptions are certainly reminiscent of cut-off distance (radii) and of many inter-atomic formulations such as the Van der Waals force or the Axilrod-Teller-Muto potential \cite{axilrod1943interaction,muto1943force}.

{
Assuming a vanishing mapping $\psi_{\omega}(D)$ is given and applied element-wise, several approaches have been proposed in order to extend the self-attention mechanism.
}
One popular extension is performed inside the softmax function as in \cite{wang2020axial} such that 
\begin{equation}
\begin{aligned}
\label{transformer_att_aa}
\varphi_{\omega}(Q,K_j,D)=\text{Softmax}\bigg(\frac{QK^{T}}{\sqrt{d}}+\psi_{\omega}(D)\bigg),
\end{aligned}
\end{equation}
This extension has the ability to fulfill the \emph{relevance assumption} requirement if the function $\psi_{\omega}(D)$, because of the softmax function, assigns negative values to distant atoms (at the limit $\lim_{d\rightarrow \infty} \psi_{\omega}(d)=-\infty$). However, such a requirement can be hard to design or even to learn with parameterized representation.
Another option is to extend the transformer outside of the normalization such that 
\begin{equation}
\begin{aligned}
\label{transformer_att_bb}
\varphi_{\omega}(Q,K_j,D)=\text{Softmax}\bigg(\frac{QK^{T}}{\sqrt{d}}\bigg)+\psi_{\omega}(D).
\end{aligned}
\end{equation}
In this vein and concurrent to our work, \cite{maziarka2020molecule} suggested MAT, a transformer architecture where the self-attention mechanism is defined as  $\psi_{\omega}(D)=\omega \text{exp}(-D)$, with hyper-parameter $\omega \in \mathbb{R}$. This approach clearly fails in fulfilling the assumption presented above since distant atoms still impact the self-attention via the softmax similarity.

We propose to directly multiply the distance relation by the similarity tensor as follows 
\begin{equation}
\begin{aligned}
\label{transformer_att_cc}
  \vspace{-0.5em}
\varphi_{\omega}(Q,K_j,D)=\text{Softmax}\bigg(\frac{QK^{T}}{\sqrt{d}}\bigg)\odot \psi_{\omega}(D),
\end{aligned}
\end{equation}
with $\odot$ denoting the Hadamard product.
This way, the interatomic distance has a direct \emph{gating} impact on the pairwise atomic  contribution obtained from the embedding.
Such representation is especially important for the augmentation scheme we present later.

Other multiplicative alternatives are possible, such as transferring the Hadamard product inside the softmax \cite{wu20213d,fuchs2020se,kwak2021geometry}. 
However there is a computationally demanding and hard to train need to couple the input of $\psi$ with the similarity tensor values in order to ensure the desired values of the softmax function, i.e., in that case $\phi_{\omega}(Q_{i}, K_{j}, D_{ij}) \rightarrow 0$ for large $D_{ij}$ implies that $\psi_{\omega}(D_{ij})\rightarrow \pm \infty$ and that $Q_{i}^{T}K_{j}$ and $\psi_{\omega}(D_{ij})$ are of opposite signs.)

\subsection{Learning the Graph Geometry}
\label{sec:metric}

Many methods have struggled to model the interaction function $\psi_{\omega}$. 
From force field methods to the recent learning-based approach, there is a need to empirically redefine the Euclidean pairwise distance in order to satisfy physical experimentation and/or performance. \\
Many handcrafted methods adopt molecular mechanics approximations of un/bonded interactions (e.g. stretch, bending, electrostatic or Van der Waals energies 
\cite{christensen2020fchl})
where molecular properties are obtained via the modification of the interatomic distance, generally using exponential mapping of the distance.\\
Neural network-based methods adopt a similar approach, where atomic embedding is obtained by modifying the pairwise distance using various learned handcrafted mollifiers, Gaussian radial basis functions, or complex basis in the corresponding function space \cite{schutt2018schnet,klicpera2020directional,qiao2020orbnet}.\\
One of the most critical hyper-parameters present in every method is the cut-off distance parameter, connecting only atoms which lie within the cut-off sphere.
This hyper-parameter may also change according to the property to be predicted \cite{schutt2018schnetpack}.

{
\color{black}
Here we propose to learn the pairwise metric  \emph{and} the molecule connectivity at each level of the Transformer. 
The transformation of the Euclidean distance coupled with the self-attention mechanism suggested above allows us to directly optimize the inter-atomic representation as well as the cut-off distance according to the prediction objective, removing cumbersome hyper-parameterization, and allowing \emph{soft and differentiable} construction of the adjacency matrix learned in a self-adaptive fashion.
The similarity function is now simply given by 
\begin{equation}
\begin{aligned}
\label{transformer_att_dd}
\varphi_{\omega}(Q,K,D)=\text{Softmax}\bigg(\frac{QK^{T}}{\sqrt{d}}\bigg)\odot \psi_{\omega}(D^{-1})^{2},
\end{aligned}
  \vspace{-0.5em}
\end{equation}
where $\psi_{\omega}:\mathbb{R}\rightarrow \mathbb{R}$ is an element-wise learnable function.
}

We parameterize $\psi_{\omega}$ as a shallow, fully connected neural network, and we further enforce the  positiveness of the new similarity map by squaring the filtered distances.
Transforming the (element-wise) inverse of the distance $D^{-1}$ instead of $D$ speeds up the training since $\psi_{\omega}$ transforms an already vanishing function (i.e. the multiplicative inverse function).

{\color{black}
In contrast with existing methods that use envelopes \cite{unke2019physnet,klicpera2020directional} or distance mollifiers \cite{schutt2018schnet}, as well as cut-off parameters, all inducing a \emph{handcrafted} graph connectivity, we optimally unify the learning of the pairwise metric and of the graph adjacency.
}

We present the learned metrics for several of the Transformer blocks in Figure \ref{fig:filters}.
As can be seen, the metric obtained is both more complex and more abstract than monotonic or Gaussian functions used in previous works \cite{schutt2018schnet,unke2019physnet,maziarka2020molecule}.
It is interesting to notice that some of the obtained cut-off distances lie around values empirically set in other works ($2-6 \angstrom $).
The diversified filter bank demonstrates the \emph{dynamic} graph connectivity the network learns via the induced masking of the similarity map.
 
 \subsection{Regularization via Molecule Augmentation}
 \label{sec:reg_sec}
Transformers are generally extremely large and over-parameterized models.
Dropout layers commonly used in Transformers in order to avoid overfitting cannot be used in our setting, because of the permutation-invariance requirement as described in \cite{lee2019set}.
One of the most efficient techniques for reducing overfitting is data augmentation.
However it is not straightforward to augment molecular data (elements of a set), especially not for regression tasks since modification of one atom type or its spatial positions has unpredictable effects on molecular properties.

While many augmentation methods modify each datum, Mixup strategies \cite{zhang2017mixup} intend to create new data samples from pairs (or more) of data. 
Here, following our initial \emph{distance relevance assumption}, we propose to extend the Mixup idea to molecules where a new data sample is obtained by creating a new system of two molecules positioned far apart.

In our mixup scheme, we constrain the property of the new system to be the sum of the properties of the two sub-molecules, even if the predicted property is intensive, i.e., unlike extensive properties it is not physically additive. The sum of the contribution of each atom in the output module allows the model to learn to disentangle the two distanced sub-systems and reduce overfitting.

Formally, for a given property $p$ and given two centered molecules $M_{i},M_{j}$ such that $M_k=\{z^{k},X^{k}\}$ we have 
\begin{equation}
\begin{aligned}
\label{augment}
&\tilde{X}^{j}\coloneqq \{R\cdot x_{1}+T,\dots,R\cdot x_{N_{j}}+T\} \\
&M_{ij} = M_{i} \cup M_{j} \coloneqq \Big\{z=\{z^{i},z^{j}\},X=\{X^{i},\tilde{X}^{j}\}\Big\},
\end{aligned}
\vspace{-0.5em}
\end{equation}
 with a rotation matrix $R\in \text{SO}(3)$, and $T=t\cdot \mathbb{1}, t \in \mathbb{R}$ a spatial translation vector ensuring large enough distance between the molecules so that the interaction is null or negligible (e.g. $t>10^3 \angstrom$).  Thus, based on inter-atomic distances, we want our model to be able to differentiate the two systems such that the target properties $p(M_{i})$ and $p(M_{j})$ sum, thus 
\begin{equation}
\begin{aligned}
\label{augment2}
p(M_{ij}) = p(M_{i} \cup M_{j}) = p(M_{i})+p(M_{j}).
\end{aligned}
\vspace{-0.5em}
\end{equation}

\begin{figure}[t]
  \begin{minipage}[c]{0.4\columnwidth}
  \centering
\includegraphics[trim={8 10 12 5},clip,width=1\columnwidth]{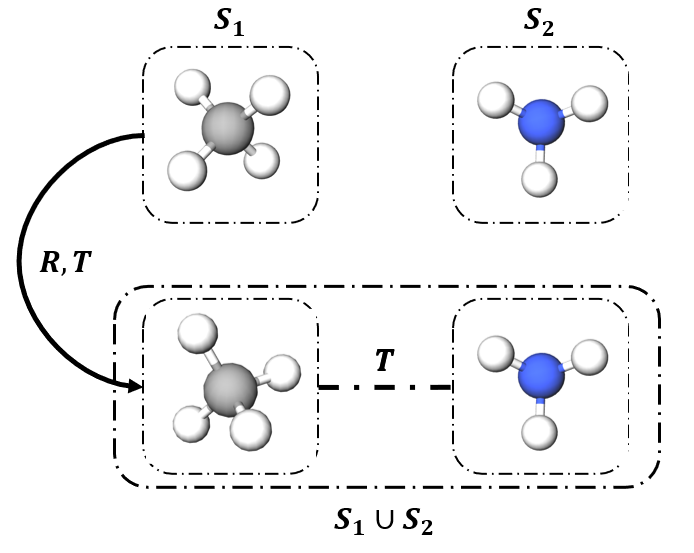}
  \end{minipage}\hfill
    \begin{minipage}[c]{0.5\columnwidth}
\caption{The proposed augmentation scheme. Given two molecular systems $S_1$ and $S_2$, we create a new system $S_1\cup S_2$ by performing random rigid transformation on the system $S_1$. }
\label{fig:augment}
  \end{minipage}
  \vspace{-1.5em}
\end{figure}

An illustration of the augmentation scheme is given in Figure \ref{fig:augment}. We note that 
methods relying on cut-off distances cannot use this augmentation strategy in a straightforward manner since the cut-off parameters would automatically separate the two molecules. Also, this approach can be extended to more than just pairs of atoms. 
The main drawback of this method is an increase in training time since the created molecules can be up to twice as large as the largest molecule of the dataset. 
However, this augmentation method enables a significant improvement of generalization, even on relatively small datasets, as demonstrated in the Discussion section. 

\begin{table*}[t!]
\parbox{0.67\textwidth}{
\caption{MAE on QM9. Best in bold and second underlined.\\
\textcolor{black}{Transformer based architectures are to the right of the vertical line.}}
  \vspace{-0.5em}
\label{table:qm9}
\scalebox{0.85}{
 \begin{tabular}{@{}c@{~}c@{~}c@{~}c@{~}c@{~}c@{~}c@{~}|@{~}c@{~}c@{~}c@{~}c@{~}c@{}}
 \toprule
 Target & Unit & Schnet & Physnet & MGCN & Cormorant & DimeNet & SE(3)-T& R-MAT & 3D-T & GeoT & \textbf{Our}\\
 \midrule
 $\mu$ & D & 0.0330 & 0.0529 & 0.0560 & 0.038 & \underline{0.0286} & {0.051}& {0.110}& {0.045}& {0.0297*}& \textbf{0.0264} \\
 $\alpha$ & $a_{0}^{3}$ & 0.235 &  0.0615 & 0.085 & 0.0681 & \textbf{0.0469} & {0.142}& {0.082}& {0.086}& {0.052}& \underline{0.051}\\
 $\epsilon_{\text{HOMO}}$ & meV &  41.0 & 32.9 &  42.1 & 34 & 27.8 & {35}& {31}& \textbf{21}& \underline{25}*& {27.5}\\
 $\epsilon_{\text{LUMO}}$ & meV & 34.0 & 24.7 &  57.4 &  38 & \textbf{19.7}& {33} & {29}& {26}& {20.2}*& \underline{20.4}\\
 $\Delta_{\epsilon}$ & meV &  63.0 & 42.5 & 64.2 & 38 & \textbf{34.8} & {53}& {48}& {39}& {43}& \underline{36.1}\\
 $\langle R^{2}\rangle$ & $a_{o}^{3}$ & \textbf{0.073} & 0.765 & 0.110 & 0.961 & 0.331& {-} & {0.676}& {-}& {0.30}& \underline{0.157}\\
 ZPVE & meV &  1.70 & 1.39 & 1.12 & 2.03 & \underline{1.29} & {-}& {2.23}& {-}& {1.7}*& \textbf{1.24}\\
 $U_{0}$ & meV &  14.0 & 8.15 & 12.9 & 22 & \underline{8.02} & {-}& {12}& {-}& {11.1}& \textbf{7.35}\\
 $U$ & meV &  19.0 &  8.34 & 14.4 & 21 & \underline{7.89} & {-}& {10}& {-}& {11.7}& \textbf{7.55}\\
 $H$ & meV & 14.0 & 8.42 & 14.6 & 21 & \underline{8.11} & {-}& {10}& {-}& {11.3}& \textbf{7.73}\\
 $G$ & meV & 14.0 & 9.40 & 16.2 & 20 & \underline{8.98} & {-}& {10}& {-}& {11.7}& \textbf{8.21}\\
 $c_{v}$ & $\frac{\text{cal}}{\text{mol K}}$ & 0.0330 &  0.0280 & 0.0380 &  \underline{0.026} & \textbf{0.0249}& {0.054} & 0.036& {-}& {0.0276}& {0.0280}\\
  \bottomrule
  \vspace{-4.5em}
\end{tabular}
}%
}%
\hfill
\parbox{.32\textwidth}{
\caption{{MAE on MD17 forces using 1000 training samples. 
}}%
\label{table:md17}
\scalebox{0.87}{
 \begin{tabular}{@{}l@{~}c@{~}c|@{~}c@{~}c@{}}
 \toprule
 Target & Schnet  & DimeNet & GeoT & \textbf{Our}\\
 \midrule
 Aspirin & 1.35 & \underline{0.499} & 0.85  & \textbf{0.451}\\
 Benzene & 0.31 & \underline{0.187} & \textbf{0.135}  & 0.28\\
 Ethanol & 0.39 & 0.230 & \underline{0.225}  & \textbf{0.212}\\
 Malonaldehyde & 0.66 & \underline{0.383} & 0.402  & \textbf{0.369}\\
 Naphthalene & 0.58 & \textbf{0.215} & -  & \underline{0.44}\\
 Salicylic acid & 0.85 & \textbf{0.374} & -  & \textbf{0.372}\\
 Toluene & 0.57 & \textbf{0.216} & 0.328  & \underline{0.24}\\
 Uracil & 0.56 & \textbf{0.301} & -  & \textbf{0.301}\\
  \bottomrule
  \\
  \\
  \\
  \\
\end{tabular}
}
}
\end{table*}

\section{Experiments}
\label{sec:experiments}
The experimental setup including  the architecture details and the training procedure is provided in the Appendix.


\subsection{QM9}
The popular QM9 dataset \cite{ramakrishnan2014quantum} contains $130,831$ molecules with up to 9 atoms of the type C,N,O, and F saturated with hydrogen atoms in their equilibrium geometries, with chemical properties computed with DFT solvers.
Following previous work, we split the dataset to $110,000$, $10,000$ and $10,831$ molecules for the training, validation and testing sets respectively.
We use the atomization energy for $U0, U, H,$ and $G$.

In Table \ref{table:qm9} we report the mean absolute error (MAE) on all QM9 targets and compare it to the state-of-the-art
models SchNet \cite{schutt2018schnet}, PhysNet \cite{unke2019physnet}, MGCN \cite{lu2019molecular}, Cormorant \cite{anderson2019cormorant}, and  DimeNet \cite{klicpera2020directional}. We also compare with concurrent molecular data Transformers, R-MAT\cite{maziarka2021relative},
3D-T \cite{wu20213d}, SE(3)-T \cite{fuchs2020se} and GeoT \cite{kwak2021geometry}. 

The method outperforms or is similar to state-of-the-art methods for most of the properties and surpasses Dimenet by $5.22\%$ average performance ratio, and other Transformers methods by large margins.
In contrast to other works \cite{schutt2018schnetpack,klicpera2020fast}, the same model and training procedure were applied for all properties.
The proposed framework does not require tuning of physical hyper-parameters or chemical approximations, making it a true end-to-end framework. 

\textcolor{black}{
The recent SOTA work DimeNet++ \cite{klicpera2020fast} substantially outperforms DimeNet (by $9\%$ in average) with careful initialization and architectural modifications, and outperforms our framework by $0.78\%$ (average performance ratio).
We believe that a similar thorough architecture search may have similar impact on the proposed approach, depending on available computational resources since Transformers are computationally intensive. 
}
It is also important to notice that the proposed method does not take into account computationally heavy bond angles between triplets of atoms (i.e. potentially inducing cubic complexity) as in many recent frameworks.

\subsection{MD17}
We use MD17 \cite{chmiela2017machine} to test model performance in molecular dynamics simulations. 
The goal of this benchmark is to predict, for eight
small organic molecules, the Cartesian atomic forces acting on each atom due to the overall potential energy.
A separate model is to be trained for each molecule, in order to provide accurate individual predictions. 
We test our model in the challenging 1000 training samples setting \cite{chmiela2018towards}. 
The original training objective is extended to molecular dynamics predictions by backpropagating to the atom coordinates $X$ as follows
\begin{equation}
\begin{aligned}
\label{md17_eq}
\vspace{-0.5em}
\mathcal{L}= ||f_{\theta}(z,D)-p||_{1}+\frac{\rho}{3}{\|-\partial_{X}f_{\theta}(z,X)-F\|_{1}},
\end{aligned}
\vspace{-0.25em}
\end{equation}
where $F$ denotes the nuclear three-dimensional Cartesian forces to be predicted and $\rho$ is the forces' loss coefficient. In our experiments $\rho$ is set to $10^{3}$, the augmentation scheme is extended straightforwardly to forces (i.e. concatenation) due to their translation invariance, and ReLU activations are replaced with GELU non-linearities to enforce twice continuous differentiability.
As shown in Table \ref{table:md17} our framework sets or reaches SOTA performances even when the training size remains extremely small, a challenging setting for Transformer models. 
Also, it demonstrates our method's flexibility and its ability to generalize to other tasks and datasets.

%

\subsection{Discussion}
\label{sec:discussions}
In this section, we study the contributions of our method.
\subsubsection{Geometric Self-attention Analysis}

\begin{figure}[ht!]
\centering
\begin{subfigure}{0.47\columnwidth}
  \centering
  \includegraphics[trim={35 100 25 90},clip, width=1\linewidth]{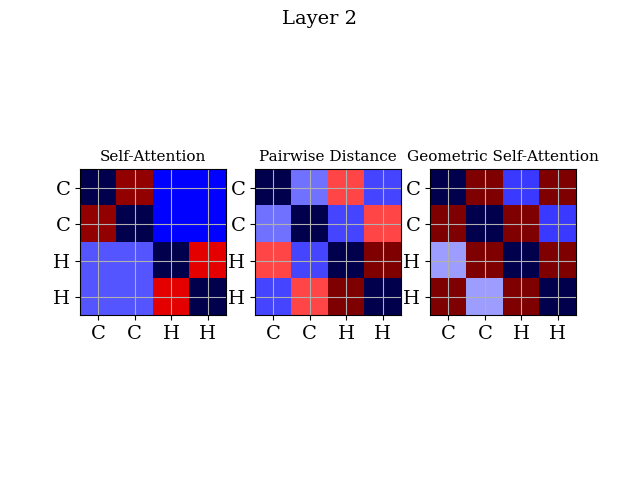}
\end{subfigure}%
\rulesep
\begin{subfigure}{0.47\columnwidth}
  \centering
  \includegraphics[trim={35 100 25 90},clip, width=1\linewidth]{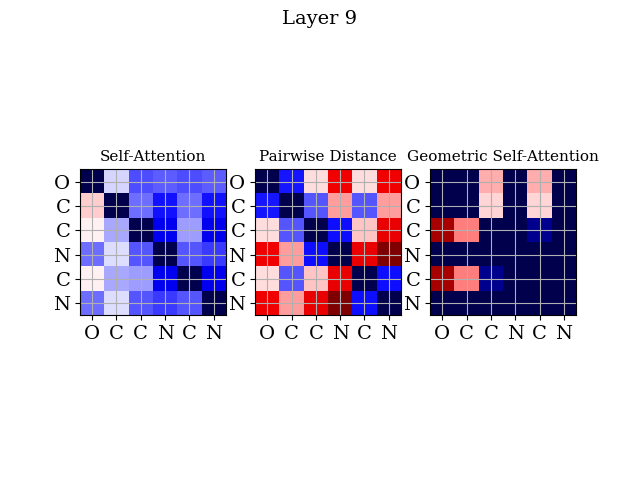}
\end{subfigure}%
\caption{Regular self-attention similarity map (left), the pairwise distance matrix (center), and the resulting geometric self-attention similarity map (right) of two different molecules (left and right) at layers 2 and 9 respectively. 
The similarity maps are averaged over the dimensions of the self-attention heads.}
\label{fig:SA_map}
\end{figure}

Figure \ref{fig:SA_map}, presents typical impact of the proposed self-attention mechanism on the similarity map. As can be seen, different filters applied to the pairwise distance have a major impact on the similarity matrix and redefine the adjacency matrix at each level of the network.
One can observe the drastic impact (color flipping) of the learned metric on the similarity map.

\subsubsection{Comparison and Ablation Studies}
\label{sec:aug_comparison}
\begin{figure}[th!]
\centering
  \centering
  \includegraphics[trim={0 0 40 0},clip, width=0.49\linewidth]{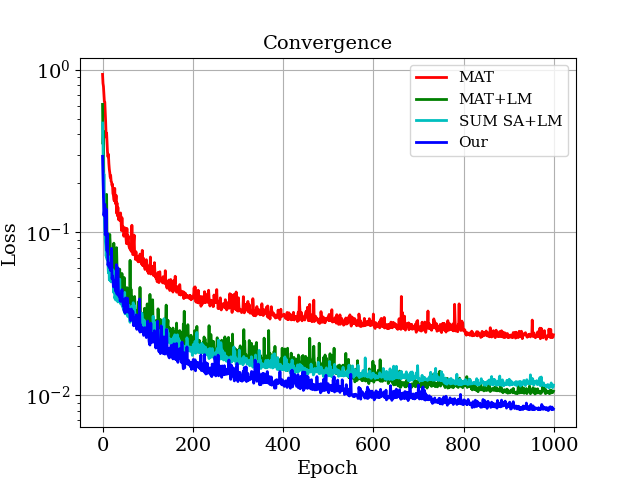}
    \includegraphics[trim={0 0 40 0},clip, width=0.49\linewidth]{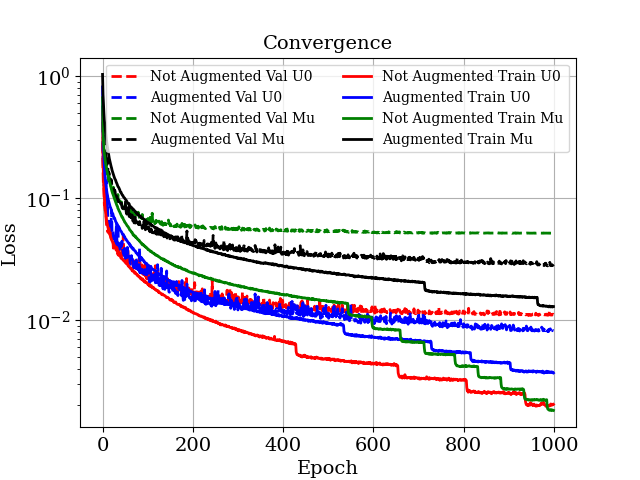}
\vspace{-0.5em}
  \caption{Left: Comparison of validation MAE losses between MAT, MAT with the proposed learned metric (MAT+LM), the summed self-attention mechanism with learned metric (SUM SA +LM), and our method.
  Right: The impact of  data augmentation on the MAE convergence and generalization of the network (val=validation) for $U0$ and $\mu$. 
  {
  Continuous lines are consistently below their corresponding dashed ones.}
  }
\label{fig:self_att1}
\vspace{-0.5em}
\end{figure}

Our ablation study compares typical impact of the different self-attention modules.
We present the convergence curves of our method from Eq. (\ref{transformer_att_dd}), the concurrent Transformers based MAT method \cite{maziarka2020molecule} from Eq. (\ref{transformer_att_bb}), the MAT method with a learned metric $\psi_\omega$, and the sum self-attention from Eq. (\ref{transformer_att_aa}) \cite{wang2020axial} with learned metric; we refer to the last method as SUM SA.
The compared networks and the training procedure are exactly the same, except for the aforementioned self-attention equation itself and the distance mapping module.
The results are presented in Figure \ref{fig:self_att1} (left). As can be seen, the MAT architecture presents the worse convergence, while the metric learning module significantly improves the performance. 
The proposed self-attention mechanism greatly surpasses all other architectures.

Finally, we present the typical effect of the data augmentation procedure on the generalization of the network during training. Figure \ref{fig:self_att1} (right) presents the impact of the augmentation on the MAE convergence of the model for both an extensive and intensive property, namely $U0$ and $\mu$. 
{\color{black} As can be seen, in both cases, when applying augmentation, the generalization gap is extremely reduced while the validation loss is much lower. The gain for the property $\mu$ is $175\%$ in terms of testing MAE and $75\%$ for $U0$. 
}

\section{Conclusion}
We introduce a new Transformer architecture and a training scheme for molecular predictions. The proposed model allows effective representation of interactions based solely on pairwise distances.
The graph geometry and connectivity can be learned in a soft fashion by the network via the geometric self-attention module. We further propose a molecular data augmentation procedure based on mixing strategies, which leads to a clear improvement in generalization for the proposed scheme.
Our results indicate that our method is the {{first Transformer}}, as far as we can ascertain, that is able to model molecular data successfully without requiring any assumptions on the underlying physical model or involving complex geometric priors.
We believe the advent of new datasets and new transformer architectures will allow the development of more efficient models also less prone to overfitting, making it a tool of predilection for the analysis of molecular data.

\section*{Acknowledgments}
This project has received funding from the European Research Council (ERC) under the European Union's Horizon 2020 research and innovation programme (grant ERC CoG 725974). The contribution of the first author is part of a PhD thesis research conducted at Tel Aviv University.
\bibliographystyle{named}
\bibliography{references2.bib}
\appendix
\section{Experimental Setup}

We used ten encoding blocks with an embedding size of $d=512$ and with a dimension of $2048$ for the inner layer of the feedforward network.
The feedforward network is composed of GEGLU layers \cite{shazeer2020glu} and layer normalization is set in pre-layer norm setting as in \cite{opennmt,xiong2020layer}.
The contribution of the atom itself (i.e. $\varphi_{\omega}(Q_{i},K_{i},D_{ii})$) is omitted (masked) in the self-attention mechanism.
The metric feed-forward module is a fully connected neural network with a 50-dimensional hidden layer and ReLU non-linearities in order to simulate distance thresholding, expanded to all the heads of the self-attention module.
The geometric positional encoder is a fully connected network with one 1024 dimensional layer coupled with GELU non-linearity.
The output module is a linear layer.
As other methods, we train our model once for every target.

The Adam optimizer \cite{kingma2014adam} is used with 32 molecules per mini-batch for the QM9 dataset and 4 molecules only for the MD17 dataset.
Half of each shuffled batch is augmented using the procedure presented in the previous section with translation scalar $t$ set to $10^4$.
We initialized the learning rate to $10^{-4}$ coupled with a plateau region decay scheduler with ratio 0.8 down to a $10^{-6}$ threshold.
No warmup has been employed \cite{xiong2020layer} and the $L_{1}$ loss is used as the objective metric (Mean Absolute Error). 
The model has been implemented upon the Schnetpack framework \cite{schutt2018schnetpack}.

\end{document}